\documentclass[sigconf]{acmart}

\setcopyright{cc}
\setcctype{by}
\copyrightyear{2026}
\acmYear{2026}
\acmConference[KDD '26]{Proceedings of the 32nd ACM SIGKDD Conference on Knowledge Discovery and Data Mining V.2}{August 09--13, 2026}{Jeju Island, Republic of Korea}
\acmBooktitle{Proceedings of the 32nd ACM SIGKDD Conference on Knowledge Discovery and Data Mining V.2 (KDD '26), August 09--13, 2026, Jeju Island, Republic of Korea}
\acmDOI{10.1145/3770855.3817531}
\acmISBN{979-8-4007-2259-2/2026/08}

\graphicspath{{figures/}}

\usepackage{multirow}
\usepackage{booktabs}
\usepackage{amsmath}
\usepackage{makecell}
\usepackage{enumitem}
\usepackage{adjustbox}
\usepackage{tabularx}

\begin{document}

\title{UniHam: A Large-Scale SOC-Complete Dataset and Benchmark for Hamiltonian Learning in Materials}

\author{Yuewen Huang}
\affiliation{%
  \institution{Sun Yat-Sen University}
  \city{Guangzhou}
  \country{China}
}
\email{huangyw66@mail2.sysu.edu.cn}

\author{Pin Chen}
\authornote{Corresponding author.}
\affiliation{%
  \institution{Sun Yat-Sen University}
  \city{Guangzhou}
  \country{China}
}
\email{chenp85@mail.sysu.edu.cn}

\author{Yutong Lu}
\authornotemark[1]
\affiliation{%
  \institution{Sun Yat-Sen University}
  \city{Guangzhou}
  \country{China}
}
\email{luyutong@mail.sysu.edu.cn}

\renewcommand{\shortauthors}{Yuewen Huang, Pin Chen, and Yutong Lu}

\begin{abstract}
Accurate prediction of electronic Hamiltonians would enable broad property inference while avoiding the high computational cost of Density Functional Theory (DFT). However, progress toward general-purpose materials foundation models is limited by a data bottleneck: existing Hamiltonian datasets are typically small, lack structural diversity, and often omit essential relativistic physics such as spin--orbit coupling (SOC). We therefore construct UniHam, a large-scale Hamiltonian dataset and benchmark suite comprising 100,000+ DFT-computed complex-valued Hermitian Hamiltonians with full SOC, covering 72 elements and a wide range of crystal geometries and symmetries (spanning diverse lattice types and space-group families). Building on UniHam, we benchmark two representative state-of-the-art models under a standardized protocol and introduce complementary evaluation metrics that jointly assess three dimensions: (i)~Hamiltonian reconstruction accuracy, (ii)~out-of-distribution (OOD) generalization across composition/symmetry shifts, and (iii)~the ability to support downstream property prediction from the predicted Hamiltonians. Experiments on UniHam demonstrate that the proposed benchmark and metrics effectively differentiate model capabilities, revealing intrinsic SOC- and element-dependent failure modes, large variations in compositional OOD robustness, and the necessity of spectral-level evaluation to assess whether Hamiltonian predictions reliably support downstream electronic-structure properties. Overall, UniHam provides a reproducible, SOC-complete benchmark that can sharpen model comparisons and accelerate the development of next-generation foundation models for quantum materials.
\end{abstract}



\maketitle

\section{Introduction}

\begin{figure*}[tbp]
    \centering
    \Description{Overview of the UniHam Benchmark Framework showing three stages: data generation, model training, and comprehensive evaluation.}
    \includegraphics[width=0.75\textwidth]{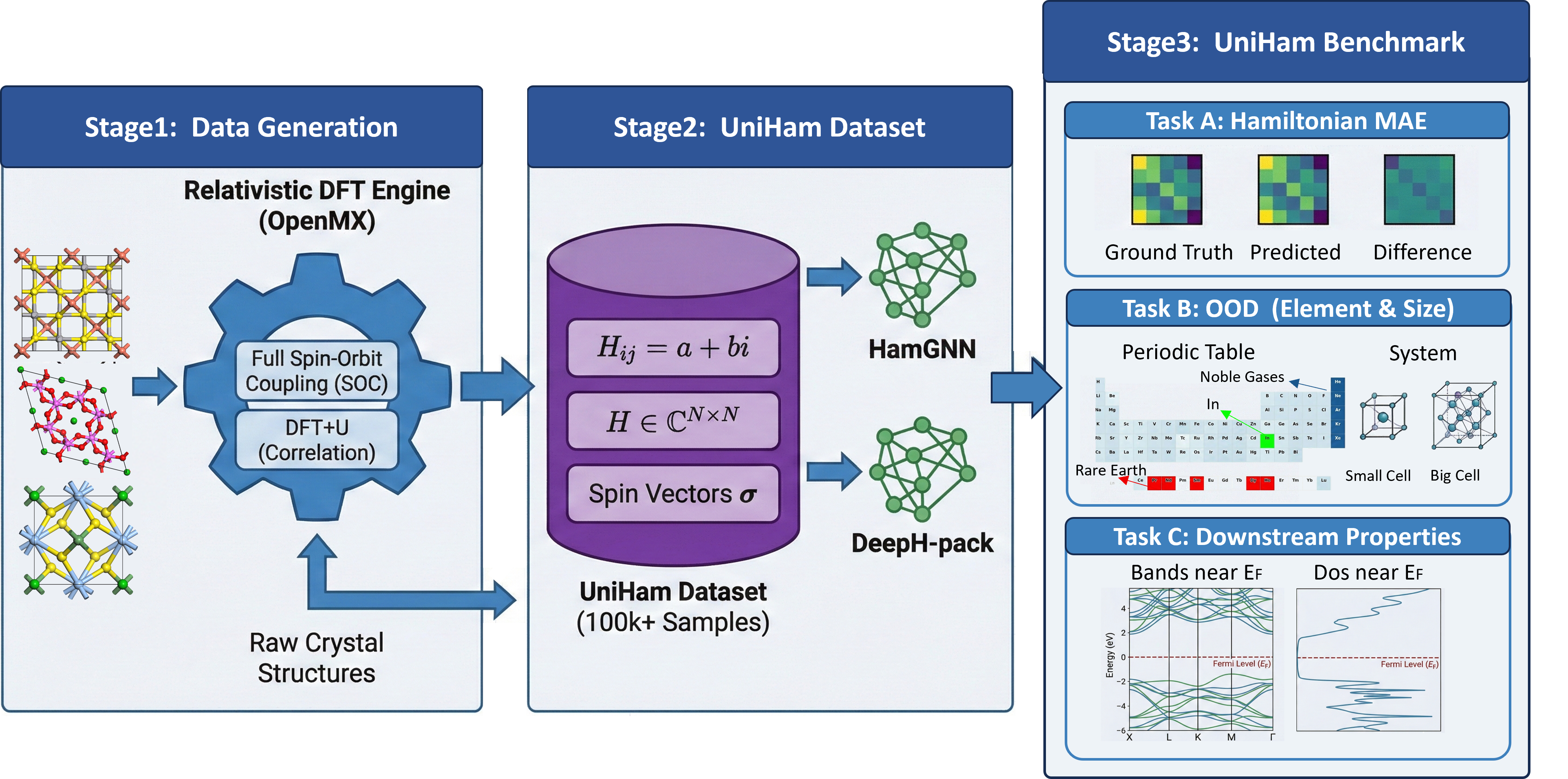}
    \caption{Overview of the UniHam Benchmark Framework. The pipeline consists of three stages: (1) Data Generation: Constructing high-quality Hamiltonian labels using relativistic DFT (OpenMX) with full Spin-Orbit Coupling (SOC) and DFT+U corrections. (2) Benchmarking: Training representative models (HamGNN, DeepH) on the UniHam dataset containing 100k+ complex-valued Hamiltonians. (3) Comprehensive Evaluation: Assessing model performance across three dimensions: Hamiltonian Reconstruction (MAE), Out-of-Distribution (OOD) Generalization (testing both unseen elements and varying atom counts), and Downstream Physical Properties (Bandgaps, Band structures, and DOS near the Fermi level).}
    \label{fig:first_pic}
\end{figure*}

First-principles electronic-structure calculations based on Density Functional Theory (DFT)~\cite{hohenberg1964inhomogeneous,kohn1965self} are central to computational materials design~\cite{jain2016computational}, yet the cubic scaling of self-consistent Kohn--Sham solvers limits large-scale and high-throughput applications. Machine-learning surrogates that bypass these expensive computations have therefore become an active research direction~\cite{behler2007generalized,zhang2018deep,huang2023central}. Most existing surrogates target scalar quantities such as total energy or band gaps; a more informative target is the \textbf{electronic Hamiltonian matrix} itself~\cite{hegde2017machine,marzari2021electronic}. Because the Hamiltonian encodes the complete single-particle electronic structure, band structures, densities of states (DOS), and other spectral observables can all be obtained from a single diagonalization without re-running self-consistent field (SCF) iterations~\cite{tang2025deep}. Recent equivariant graph neural networks have shown that structure-to-Hamiltonian mappings are learnable for selected material families~\cite{schutt2019unifying,li2022deep,wang2024universal,li2025critical}.

Scaling Hamiltonian learning models toward general-purpose applicability is fundamentally bottlenecked by the availability of suitable \emph{training and evaluation data}. Compared to energies or forces, a Hamiltonian label is a high-dimensional, symmetry-constrained matrix whose accurate construction requires fully self-consistent electronic-structure calculations, making large-scale curation substantially more expensive. Moreover, existing collections of Hamiltonian data are typically limited in at least one of the following aspects: \emph{scale}, \emph{chemical diversity}, \emph{public accessibility}, or \emph{physical completeness}.

A particularly severe limitation is the lack of systematic inclusion of \textbf{spin--orbit coupling (SOC)}. Without SOC, Hamiltonians are restricted to real-valued forms, which are inadequate for heavy-element, magnetic, and topological materials where relativistic effects are essential~\cite{galitski2013spin}. Incorporating SOC introduces complex-valued spin-mixing terms and effectively doubles the matrix dimension, further amplifying both data-generation cost and learning difficulty. Consequently, the community currently lacks an \emph{open, large-scale, and SOC-complete} Hamiltonian dataset that can support reproducible benchmarking, scaling analysis, and robust out-of-distribution evaluation across broad materials space.

We present \textbf{UniHam}, an open, large-scale dataset and benchmark for crystal Hamiltonian prediction with full SOC. UniHam contains \textbf{103,121} structures covering \textbf{72} elements and 225 of 230 crystallographic space groups, with every Hamiltonian computed under non-collinear spin and full relativistic SOC using OpenMX~\cite{ozaki2003variationally,ozaki2004numerical}, yielding \textbf{complex-valued Hermitian} matrices that capture spin--orbit and spin-mixing interactions. The dataset construction consumed approximately \textbf{3 million CPU core-hours} on national supercomputing clusters. Our main contributions are:
\begin{itemize}
    \item We construct UniHam, a large-scale dataset of 100K+ DFT-computed complex-valued Hermitian Hamiltonians with full spin--orbit coupling, spanning 72 elements and diverse crystal geometries and symmetries, substantially expanding the data foundation for learning electronic Hamiltonians in materials.

    \item We establish a standardized benchmark with complementary metrics that jointly assess Hamiltonian reconstruction accuracy, compositional and symmetry-aware out-of-distribution generalization, and the ability of predicted Hamiltonians to support downstream electronic-structure property prediction.

    \item Through extensive evaluation, we show that UniHam reveals SOC-driven and element-dependent failure modes and sharp performance gaps under compositional OOD settings, and demonstrate that incorporating spectral-level evaluation is essential for judging whether Hamiltonian accuracy translates into reliable downstream properties.

\end{itemize}

\section{Related Work}
\label{sec:related_work}
We review related work from two complementary perspectives:
(i) existing materials Hamiltonian datasets for machine-learning-driven electronic-structure modeling, and
(ii) neural architectures for predicting crystal Hamiltonians from atomic structure.
Together, these lines of research highlight a persistent gap between the growing demand for general-purpose Hamiltonian learning and the limited availability of large-scale, open, and SOC-complete benchmarks---motivating the construction of \textsc{UniHam} and the systematic evaluations presented in this work.

\subsection{Hamiltonians Dataset}
Existing materials Hamiltonian datasets differ substantially in scale, chemical coverage, relativistic physics (e.g., SOC), and---critically---\emph{public availability}. Early DeepH-style datasets were largely system-specific (e.g., graphene/C and MoS$_2$) and primarily served as internal training/evaluation corpora~\cite{li2022deep}. To illustrate a more general materials-level setting, a subsequent DeepH study curated a 12,062-structure crystal set from Materials Project~\cite{jain2013commentary} under explicit filtering rules (e.g., non-magnetic only, first four rows with Sc--Ni excluded to avoid magnetism, noble gases excluded, and unit cells capped at 150 atoms); however, this Hamiltonian dataset is not released as a public benchmark~\cite{wang2024universal}. Similarly, HamGNN reports a larger dataset of $\sim$44k structures to support universal Hamiltonian learning, but the corresponding Hamiltonian data are not publicly available~\cite{zhong2024universal}.

In contrast, \textsc{Materials-HAM-SOC} released by NextHAM provides an open SOC-inclusive benchmark with complex-valued Hamiltonians for $\sim$17k crystals spanning $60+$ elements~\cite{yin2025advancing}. Building on the need for an \emph{open} dataset that is simultaneously larger-scale, more chemically diverse, and SOC-complete, we release \textsc{UniHam}, comprising 100k\,+ crystals with full-SOC complex Hermitian Hamiltonians over 72 elements, enabling reproducible scaling studies and robust out-of-distribution evaluation.

Table~\ref{tab:dataset_comparison} provides a feature-level comparison between UniHam and prior Hamiltonian datasets. UniHam is $6\times$ larger than the next public SOC dataset, the only one covering lanthanide ($4f$) elements, and the only benchmark with systematic OOD evaluation and gauge-corrected multi-level metrics.

\begin{table}[t]
\centering
\footnotesize
\caption{Feature-level comparison between UniHam and prior Hamiltonian datasets. NR = not reported.}
\label{tab:dataset_comparison}
\resizebox{\columnwidth}{!}{%
\begin{tabular}{@{} l c c c c @{}}
\toprule
\textbf{Feature} & \textbf{UniHam (Ours)} & \textbf{Mat-HAM-SOC} & \textbf{DeepH} & \textbf{HamGNN} \\
 & & \cite{yin2025advancing} & \cite{wang2024universal} & \cite{zhong2024universal} \\
\midrule
Structures & 103,121 & $\sim$17,000 & 12,062 & $\sim$44,000 \\
Elements & 72 & 68 & $\sim$24 (per. 1--4) & NR \\
Lanthanide ($4f$) & 9 (La--Lu) & Excluded & Excluded & NR \\
Full SOC & All & All & No & No \\
Heavy elem.\ ($Z{\ge}37$) & 66.6\% & $\sim$50\% & 0\% & NR \\
Max atoms/cell & 444 & $\sim$80 & 150 & NR \\
DFT+U & Yes & NR & No & NR \\
Publicly available & Yes & Yes & No & No \\
Systematic OOD eval. & Yes & No & No & No \\
\bottomrule
\end{tabular}%
}
\end{table}

\subsection{Hamiltonians Prediction Models}
Hamiltonian prediction from atomic structure is commonly formulated as learning local Hamiltonian blocks in a localized orbital basis with symmetry-aware graph models. DeepH establishes a representative structure-to-Hamiltonian paradigm and demonstrates strong accuracy in system-specific settings~\cite{li2022deep,gong2023general,li2023deep_magnetic}. HamGNN advances transferability via O(3)-equivariant GNN designs and training objectives that improve generalization across chemistries~\cite{zhong2023transferable,zhong2024universal}. NextHAM further explores SOC-inclusive Hamiltonian learning and proposes a delta-learning formulation conditioned on an auxiliary ``zeroth-step'' Hamiltonian $H^{(0)}$~\cite{yin2025advancing}.

However, Hamiltonians are intrinsically code- and basis-dependent: different DFT software stacks and localized orbital conventions lead to representations that are not directly comparable. Since NextHAM and UniHam adopt Hamiltonians generated by different DFT packages and basis conventions, a direct quantitative comparison would therefore confound model capability with representation mismatch. Moreover, obtaining $H^{(0)}$ requires additional electronic-structure computation beyond the input structure. Accordingly, we focus our benchmark on structure-only Hamiltonian predictors and evaluate representative equivariant architectures on the open \textsc{UniHam} dataset.

\section{Problem Formulation}
\label{sec:preliminaries}

From a machine learning perspective, Hamiltonian prediction can be understood as a \textbf{structured regression problem on periodic geometric graphs}. The input is a crystal structure represented as a graph $\mathcal{G}=(\mathcal{V},\mathcal{E})$ with atoms as nodes and interatomic neighborhoods as edges; the output is a variable-size, \textbf{block-sparse, complex-valued Hermitian matrix} whose entries transform under strict SE(3)-equivariance constraints dictated by Wigner D-matrices. This formulation connects Hamiltonian learning to broader themes in geometric deep learning and equivariant representation theory, while the high-dimensional matrix output---rather than scalar or vector targets---makes it a uniquely demanding structured prediction task.

We progressively introduce (i) the learning target and its structural properties, (ii) the additional complexity induced by Spin--Orbit Coupling (SOC), and (iii) the symmetry constraints that motivate SE(3)-equivariant neural architectures.

\subsection{Kohn--Sham Hamiltonian}

Density Functional Theory (DFT) computes electronic structure by solving the Kohn--Sham eigenproblem
\begin{equation}
    \hat{H}_{\mathrm{KS}} |\psi_i\rangle = \epsilon_i |\psi_i\rangle ,
\end{equation}
where $\hat{H}_{\mathrm{KS}}$ is the Kohn--Sham Hamiltonian operator, and $(\epsilon_i, |\psi_i\rangle)$ are eigenpairs. In practice, $\hat{H}_{\mathrm{KS}}$ is represented in a finite basis, yielding a matrix $H$.

The Kohn--Sham Hamiltonian matrix $H$ is the central object of electronic-structure theory. Its eigenvalues directly give band structures, and its eigenvectors give wavefunctions. From these fundamental quantities, a wide range of physically important observables can be derived without additional first-principles calculations: band gaps determine semiconductor applicability; band curvature yields effective masses for charge transport; the density of states near the Fermi level governs conductivity and optical spectra; and the global structure of the eigenspectrum encodes topological invariants such as Chern numbers and $\mathbb{Z}_2$ indices. This makes $H$ a uniquely informative prediction target: unlike scalar properties such as total energy, a single accurate Hamiltonian enables simultaneous inference of dozens of downstream properties via low-cost post-processing, effectively bypassing the expensive self-consistent field iterations that dominate conventional DFT workflows.

We therefore cast Hamiltonian prediction as a supervised learning problem
\begin{equation}
    \Phi: \mathcal{G} \rightarrow H ,
\end{equation}
where $\mathcal{G}$ denotes a crystal structure.

We adopt a localized pseudo-atomic orbital (PAO) basis~\cite{prayitno2018implementation}, under which a crystal is naturally represented as a periodic graph $\mathcal{G}=(\mathcal{V},\mathcal{E})$ with atoms as nodes and interatomic neighborhoods as edges. The Hamiltonian becomes block-sparse: each block $H_{ij}$ corresponds to couplings between orbitals on atoms $i$ and $j$.

From a machine learning perspective, this target exhibits several important properties: (i) $H$ is \textbf{Hermitian}, satisfying $H_{ij}=H_{ji}^{\dagger}$; (ii) $H$ is \textbf{local and sparse}, since only atom pairs within a finite cutoff contribute non-zero blocks; and (iii) each block has a \textbf{fixed internal structure} determined by the orbital basis. Accordingly, learning proceeds at the block level: a neural network predicts orbital-resolved features for atoms and pairwise interactions, which are assembled into Hamiltonian blocks and aggregated to form the full matrix.

\subsection{Relativistic Regime and Spin--Orbit Coupling}

Accurate modeling of heavy elements and quantum materials requires explicit treatment of relativistic effects, most notably SOC. A concrete example is the topological insulator Bi$_2$Se$_3$: spin--orbit coupling inverts the conduction and valence bands at the $\Gamma$ point, opening a topologically non-trivial gap. Without SOC, this phase is completely invisible---the material would be incorrectly predicted as a trivial insulator. SOC also governs band splittings in heavy-element semiconductors (e.g., the $\sim$0.5\,eV splitting of the valence band maximum in PbTe), enables spin-Hall effects in spintronics, and determines the fine electronic structure of rare-earth compounds.

While prior Hamiltonian datasets provide limited or inconsistent SOC coverage, UniHam targets a SOC-complete setting. When SOC is included, the Hamiltonian acquires two essential properties:
\begin{itemize}
    \item \textbf{Complex-valued entries}, reflecting phase information and time-reversal symmetry;
    \item \textbf{Spinor degrees of freedom}, where each orbital is associated with spin-up and spin-down components, yielding intra-spin and spin-flip couplings.
\end{itemize}

Consequently, the learning target in UniHam is a \textbf{complex-valued Hermitian matrix} whose blocks encode coupled orbital--spin interactions, substantially increasing modeling difficulty compared to real-valued Hamiltonians.

\subsection{SE(3)-Equivariance and Symmetry Constraints}

Hamiltonians are constrained by the geometric symmetries of atomic structures. For a rotation $R \in SO(3)$ acting on a crystal,
\begin{equation}
    H(R \cdot \mathcal{G}) = D(R)\, H(\mathcal{G})\, D(R)^{\dagger},
\end{equation}
where $D(R)$ denotes the representation matrix associated with the orbital angular momenta of the basis functions. This relation implies that Hamiltonian prediction is inherently an \emph{equivariant} learning problem: rotating the input structure must induce a corresponding, well-defined transformation of the output matrix. SE(3)-equivariant neural networks~\cite{thomas2018tensor,geiger2022e3nn} provide a principled mechanism to encode this inductive bias by constructing features that transform according to irreducible representations of the rotation group.

Recent models such as DeepH~\cite{gong2023general} and HamGNN~\cite{zhong2024universal} adopt this paradigm, but have primarily been evaluated on small-scale or non-relativistic datasets. UniHam provides a large-scale, SOC-complete benchmark to rigorously assess equivariant architectures under the full complexity of relativistic Hamiltonian learning.

\section{The UniHam Dataset and Benchmark}
This section introduces \textsc{UniHam} and the accompanying benchmark suite for Hamiltonian learning. We first describe the dataset construction and its statistical characteristics, followed by the definition of benchmark tasks that probe reconstruction accuracy, two forms of out-of-distribution generalization, and downstream property prediction.
We then specify the evaluation metrics and the baseline models used throughout our experiments.

\subsection{Dataset}
\label{sec:dataset}
\begin{figure}[t]
    \centering
    \Description{Periodic table heatmap showing element coverage of the UniHam dataset with 72 elements color-coded by occurrence frequency.}
    \includegraphics[width=0.48\textwidth]{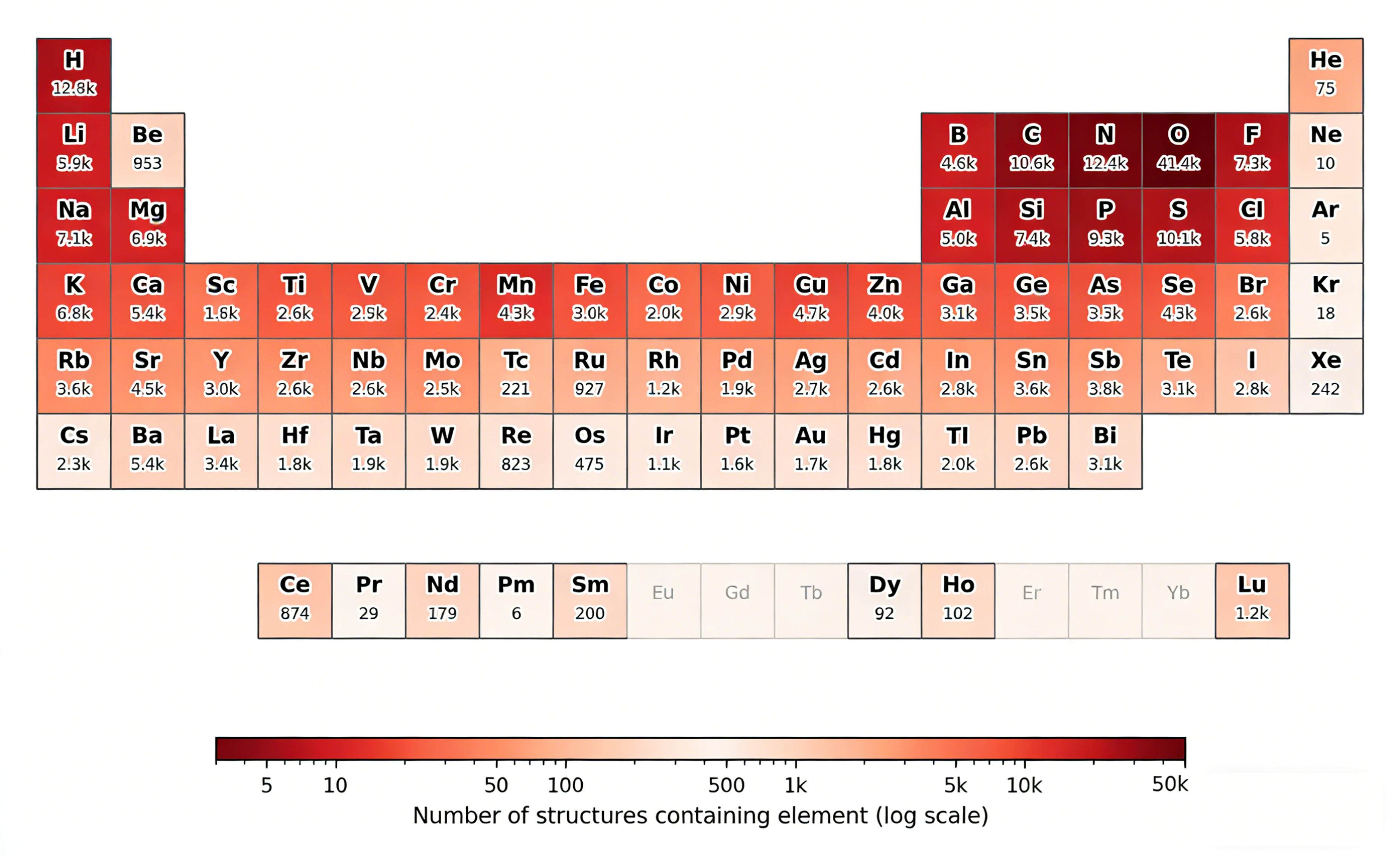}
    \caption{Periodic-table heatmap of per-element structure counts in UniHam. The 72 covered elements are color-coded by occurrence frequency.}
    \label{fig:elem_heatmap}
\end{figure}

\begin{figure}[t]
    \centering
    \Description{Histogram showing the distribution of atoms per unit cell across UniHam structures.}
    \includegraphics[width=0.38\textwidth]{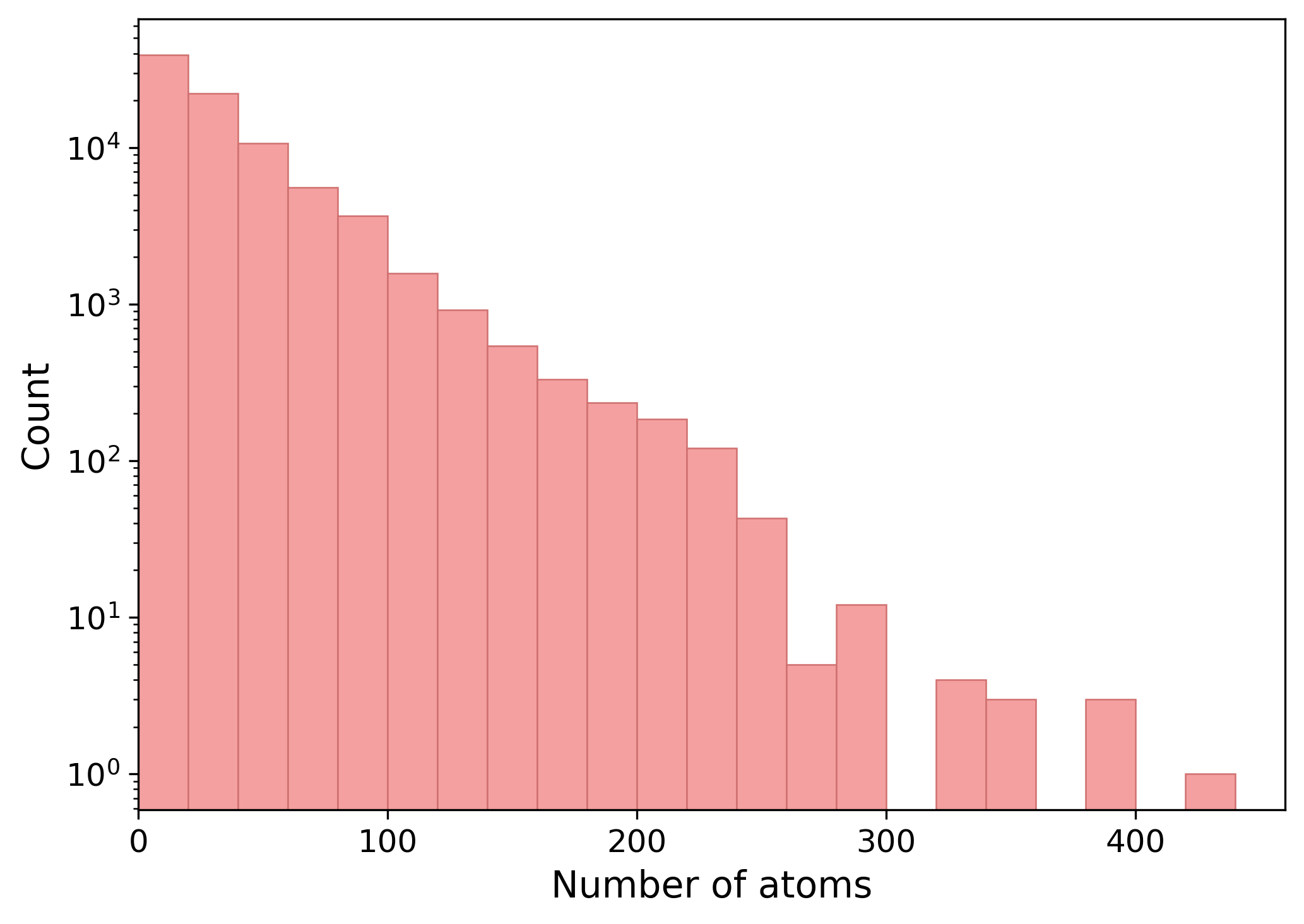}
    \caption{Distribution of the number of atoms per unit cell. Most structures contain 10--60 atoms; a long tail extends to 444.}
    \label{fig:atom_count_hist}
\end{figure}

UniHam contains \textbf{103,121} crystal structures with converged DFT Hamiltonians, drawn from ICSD~\cite{belsky2002icsd} (43.0\%), Matgen~\cite{chen2022improving} (52.5\%), and COD~\cite{graulis2012crystallography} (4.5\%). The dataset covers \textbf{72 elements} (Figure~\ref{fig:elem_heatmap}); transition metals and oxygen are the most frequently occurring, consistent with the compositional makeup of known inorganic phases. About two-thirds of the structures (66.6\%) include at least one heavy element ($Z \geq 37$), providing broad sampling of the 4$d$ and 5$d$/6$p$ regimes where SOC qualitatively affects the electronic structure. Unit-cell sizes range from 1 to 444 atoms (Figure~\ref{fig:atom_count_hist}), with a median of 20 and a mean of 30. The dataset spans \textbf{225 of 230} crystallographic space groups and all 32 point groups.

To generate high-quality Hamiltonian labels, we employ the \textbf{OpenMX} (Open source package for Material eXplorer) package~\cite{ozaki2003variationally,ozaki2004numerical}, which is based on localized pseudo-atomic orbital (PAO) basis functions. Compared with plane-wave representations, PAO bases yield compact real-space Hamiltonian matrices that are naturally compatible with graph-based learning architectures and avoid costly basis transformations. All calculations are performed within the generalized gradient approximation using the Perdew--Burke--Ernzerhof (PBE) exchange--correlation functional~\cite{perdew1996generalized}.

To ensure physical completeness, \textbf{full spin--orbit coupling (SOC)} and \textbf{non-collinear (NC) spin} are enabled for \emph{all} structures, regardless of their magnetic character. This leads to complex-valued Hamiltonians and increases the computational cost by approximately $8\times$, but is essential for describing relativistic effects in heavy elements and topological materials. Strong electron correlations in transition-metal ($3d$) and rare-earth ($4f$) elements are treated using the \textbf{DFT+U (GGA+U)} formalism~\cite{dudarev1998electron}. The effective Hubbard parameters $U_{\mathrm{eff}}=U-J$ are strictly standardized following the protocol of Chen \emph{et al.}~\cite{chen2022improving}, ensuring consistency across the entire dataset.

Self-consistent field (SCF) calculations employ a real-space grid cutoff of 350~Ry, a convergence threshold of $10^{-6}$, and a maximum of 500 SCF iterations. A robust Pulay mixing scheme is adopted with initial, minimum, and maximum mixing weights of 0.3, 0.001, and 0.4, respectively, and Pulay extrapolation activated after six iterations. An electronic temperature of 300~K is used to facilitate convergence. Brillouin-zone sampling is performed using Monkhorst--Pack $k$-point meshes with a baseline density corresponding to \textbf{2000 $k$-points per reciprocal atom (kppa)}; for structures that do not converge under this setting, the density is increased to 6000--8000 kppa. All calculations are single-step electronic evaluations without structural relaxation, ensuring that the Hamiltonians correspond exactly to the input geometries.

The full data-construction pipeline is described in Appendix~\ref{app:data_generation}. Detailed per-element analyses and additional experimental results appear in Appendix~\ref{app:detailed_experiments}.

\subsection{Tasks}
\label{sec:tasks}

To comprehensively evaluate Hamiltonian prediction models, we design four benchmark tasks that probe complementary capabilities: in-distribution reconstruction accuracy, two forms of out-of-distribution generalization, and downstream property prediction. Table~\ref{tab:tasks_metrics} summarizes the data configurations and primary evaluation metrics.

\paragraph{\textbf{Recon-id: In-Distribution Reconstruction.}}
Since SOC-complete Hamiltonians scale quadratically with the number of spinor orbitals, GPU memory consumption during training grows rapidly with system size. Even on NVIDIA H100 GPUs (80\,GB) with a batch size of 1, structures exceeding approximately 60 atoms per unit cell cause out-of-memory failures during back-propagation (see Appendix~\ref{app:impl_details} for details). We therefore restrict the training pool to structures with $\leq$60 atoms (${\sim}$90{,}000 structures) and partition them into ${\sim}$86{,}000 training, 2{,}000 validation, and 2{,}000 test structures using a \emph{stratified splitting} strategy based on rare-element balancing. Concretely, each structure is assigned to a bin defined by its rarest element (the element with the lowest global occurrence count), and structures within each bin are proportionally allocated across splits via seeded random shuffling (\texttt{seed=42}). This guarantees that even infrequent elements (e.g., Tc, Pm) appear across all splits, preventing trivially poor test-time performance on unseen chemical species. The remaining structures with $>$60 atoms are reserved for the \textsc{OOD-Size} generalization task (see below). The primary metric is gauge-corrected $\mathrm{MAE}_{H}$.

\paragraph{\textbf{OOD-Size: System-Size Generalization.}}
Models trained on the standard \textsc{Recon-id} training set (structures with $\leq$60 atoms) are tested on 4{,}328 held-out structures containing 61--444 atoms per unit cell, covering the same 72-element chemical space. This task probes whether local message-passing architectures can transfer to substantially larger unit cells without retraining.

\paragraph{\textbf{OOD-Comp: Compositional Generalization.}}
We design three scenarios of escalating chemical distance to evaluate compositional extrapolation and retrain the model:
\begin{enumerate}[leftmargin=*,nosep]
    \item \textbf{Chemical Interpolation (In, $Z{=}49$):} Indium is held out while all periodic-table neighbors (Cd, Sn, Ga, Tl) remain in training, forming a favorable interpolation setting with qualitatively identical bonding character ($sp$ hybridization).
    \item \textbf{Electronic-Structure Extrapolation (Rare Earths---Pr, Nd, Sm, Dy, Ho):} Five lanthanides with partially filled $4f$ shells ($4f^{2}$--$4f^{11}$) are held out; only La ($4f^{0}$), Ce ($4f^{1}$), and Lu ($4f^{14}$) remain, providing no coverage of the complex many-electron $4f$ manifold.
    \item \textbf{Bonding-Character Extrapolation (Noble Gases---He, Ar, Kr, Xe):} Elements with completely filled shells and fundamentally different bonding character (van der Waals / hypervalent) are held out with no chemical analogs in training.
\end{enumerate}
All structures containing any held-out element are excluded from training/validation; each test structure is assigned to exactly one scenario to avoid confounding, yielding 3{,}111 / 845 / 285 test structures for the In / RE / NG scenarios, respectively.

\paragraph{\textbf{Prop-id: Downstream Property Prediction.}}
We evaluate whether predicted Hamiltonians support downstream electronic-structure analysis by diagonalizing predicted Hamiltonians and comparing derived observables against DFT references within a physics-motivated energy window centered on the Fermi level: $[E_{\mathrm{VBM}} - 3\,\text{eV},\; E_{\mathrm{CBM}} + 3\,\text{eV}]$ for non-metals and $[-3\,\text{eV},\; +3\,\text{eV}]$ for metals. This task reports band gap error, band-energy MAE ($\mathrm{MAE}_{\mathrm{band}}$), DOS MAE ($\mathrm{MAE}_{\mathrm{DOS}}$), and eigenstate RMSE ($\mathrm{RMSE}_{\psi,\Gamma}$).

\begin{table}[t]
\centering
\caption{Task configurations and evaluation protocols.}
\label{tab:tasks_metrics}
\footnotesize
\setlength{\tabcolsep}{4pt}
\begin{tabular}{l p{3.8cm} l}
\toprule
\textbf{Task} & \textbf{Training Protocol} & \textbf{Primary Metrics} \\
\midrule
\textsc{Recon-ID}
& Random split (8:1:1)
& $\mathrm{MAE}_{H}$ \\

\textsc{OOD-Size}
& Evaluate on larger structures using \textsc{Recon-ID} model
& $\mathrm{MAE}_{H}$ \\

\textsc{OOD-Comp}
& Retrain after removing specified elements from training set
& $\mathrm{MAE}_{H}$ \\

\textsc{Prop-ID}
& Use \textsc{Recon-ID} model; diagonalize predicted Hamiltonians
& \makecell[l]{Band gap, $\mathrm{MAE}_{\mathrm{band}}$,\\$\mathrm{MAE}_{\mathrm{DOS}}$} \\
\bottomrule
\end{tabular}
\end{table}

\subsection{Metrics}
\label{sec:metrics}

We evaluate predicted Hamiltonians from both operator- and observable-level perspectives using four complementary metrics.
Since adding a global shift $\mu S$ (with overlap matrix $S$) does not change downstream physics, we report gauge-corrected errors.

\paragraph{(1) Gauge-corrected Hamiltonian MAE.}
We measure the mean absolute error over all Hamiltonian matrix elements,
\begin{equation}
\mathrm{MAE}_{H}=
\frac{\sum_{i\alpha,j\beta}\left|H^{\mathrm{pred}}_{i\alpha,j\beta}-H^{\mathrm{bm}}_{i\alpha,j\beta}-\Delta\mu_{H} S_{i\alpha,j\beta}\right|}{N_H},
\end{equation}
where $N_H$ is the total number of Hamiltonian entries.

\paragraph{(2) Band-energy MAE.}
We compute the MAE of band energies over Brillouin-zone samples and an energy window $(\varepsilon_{\min},\varepsilon_{\max})$:
\begin{equation}
\mathrm{MAE}_{\mathrm{band}}=
\frac{\int d^n\mathbf{k}\sum_{n}'\left|\varepsilon^{\mathrm{pred}}_{n}(\mathbf{k})-\varepsilon^{\mathrm{bm}}_{n}(\mathbf{k})-\Delta\mu_{\mathrm{band}}\right|}
{\int d^n\mathbf{k}\,N'_{\mathrm{band}}(\mathbf{k})}.
\end{equation}
Here $\sum_n'$ (and $N'_{\mathrm{band}}$) restrict bands within the energy window.

\paragraph{(3) DOS MAE (normalized).}
To assess spectral distribution, we compare densities of states within the same window:
\begin{equation}
\mathrm{MAE}_{\mathrm{DOS}}=
\frac{\int_{\varepsilon_{\min}}^{\varepsilon_{\max}}
\left|D^{\mathrm{pred}}(\varepsilon+\Delta\mu_{\mathrm{band}})-D^{\mathrm{bm}}(\varepsilon)\right|\,d\varepsilon}
{\int_{\varepsilon_{\min}}^{\varepsilon_{\max}} D^{\mathrm{bm}}(\varepsilon)\,d\varepsilon}.
\end{equation}

For both band energies and DOS, we additionally report the \textbf{Pearson correlation coefficient} $r$ between predicted and ground-truth values within the energy window, providing a complementary shape-similarity measure that is insensitive to uniform shifts.

\paragraph{(4) Band gap error.}
For non-metallic structures, we report the absolute difference between the predicted and DFT-reference band gaps: $\Delta E_g = |E_g^{\mathrm{pred}} - E_g^{\mathrm{bm}}|$.

\paragraph{Gauge correction.}
The gauge shifts are computed as
\begin{align}
\Delta\mu_H &= \frac{\sum_{i\alpha,j\beta}\left(H^{\mathrm{pred}}_{i\alpha,j\beta}-H^{\mathrm{bm}}_{i\alpha,j\beta}\right)S_{i\alpha,j\beta}}
{\sum_{i\alpha,j\beta} S_{i\alpha,j\beta}^2},\\
\Delta\mu_{\mathrm{band}} &= \mathrm{median}\Big\{\varepsilon^{\mathrm{pred}}_{n}(\mathbf{k})-\varepsilon^{\mathrm{bm}}_{n}(\mathbf{k}) \,\big|\,
\varepsilon_{\min}<\varepsilon^{\mathrm{bm}}_{n}(\mathbf{k})<\varepsilon_{\max}\Big\}.
\end{align}

\subsection{Baselines}
\label{sec:baselines}

We benchmark two representative state-of-the-art equivariant Hamiltonian predictors that accept crystal structure as sole input (i.e., without requiring a zeroth-step Hamiltonian $H^{(0)}$).

\paragraph{DeepH-pack.} We use the \textbf{DeepH-E3} architecture~\cite{gong2023general} from the DeepH-pack framework~\cite{li2022deep}, which achieves $E(3)$-equivariance via equivariant neural networks operating on local atomic neighborhoods. The model employs 3 message-passing blocks with 2 attention heads and a radial cutoff of 10.0\,\AA. We adopt the default loss (MAE on Hamiltonian matrix elements) and the AdamW optimizer with a ReduceOnPlateau scheduler. Since the publicly released DeepH-pack software does not natively support datasets exceeding ${\sim}$10k structures, we implement an \emph{epoch-level data sharding} strategy to enable training on the full 86k training set (details in Appendix~\ref{app:impl_details}).

\paragraph{HamGNN.} We evaluate two variants of HamGNN~\cite{zhong2024universal} to quantify the effect of spectral supervision: \textbf{HamGNN (Base)}, trained with Hamiltonian matrix element MAE loss only, and \textbf{HamGNN (FT)}, obtained by second-stage fine-tuning that additionally incorporates eigenvalue-level constraints (band-energy loss) to improve spectral fidelity. Both variants share the same architecture: 3 equivariant message-passing layers with a radial cutoff of 26.0\,Bohr (${\approx}$13.8\,\AA), 64 Bessel radial basis functions, and gate nonlinearity. Note that eigenvalue-level fine-tuning is a built-in capability of HamGNN, while DeepH-pack does not currently provide analogous functionality. We therefore report only the standard MAE-trained variant for DeepH-pack.

Both models are trained on 8 NVIDIA \textbf{H100} GPUs with FP32 precision. Full hyperparameters are provided in Appendix~\ref{app:impl_details}.

\section{Experiments}
\label{sec:experiments}
We present a comprehensive empirical evaluation of Hamiltonian prediction models on the \textsc{UniHam} benchmark.
Following the task definitions in Section~\ref{sec:tasks}, we assess in-distribution reconstruction accuracy, out-of-distribution generalization, and downstream property prediction, using the metrics described in Section~\ref{sec:metrics}.
All experiments are conducted with the baseline models introduced in Section~\ref{sec:baselines}.

\subsection{Hamiltonian Reconstruction Accuracy}

\begin{table}[h]
    \centering
    \caption{Hamiltonian reconstruction performance (MAE in meV) on the test set, decomposed by real part, imaginary part, and total.}
    \label{tab:test_split_ev}
    \begin{tabular}{l ccc}
        \toprule
        \multirow{2}{*}{\textbf{Model}} & \multicolumn{3}{c}{\textbf{Test Set MAE (meV)}} \\
        \cmidrule(lr){2-4}
         & \textbf{Real} & \textbf{Imag.} & \textbf{Total} \\
        \midrule
        DeepH-pack  & 4.49 & 0.385 & 2.44 \\
        HamGNN      & 2.96 & 0.0412 & 1.50 \\
        \bottomrule
    \end{tabular}
\end{table}

We evaluate the \textsc{Recon-id} task using the stratified split. Since SOC yields complex-valued Hamiltonians, we report MAE decomposed into real, imaginary, and total components (Table~\ref{tab:test_split_ev}). Both models achieve sub-3\,meV total MAE, confirming that equivariant architectures can accurately learn complex-valued Hamiltonians from crystal structures. HamGNN attains 1.50\,meV total MAE, outperforming DeepH-pack (2.44\,meV) by $1.6\times$, and achieves lower error on 96.5\% of test structures. Errors are dominated by the real part, while imaginary (SOC-induced spin-flip) terms are an order of magnitude smaller; notably, HamGNN exhibits a much lower imaginary-to-real MAE ratio (1.4\%) than DeepH-pack (8.6\%), indicating better resolution of fine-grained SOC couplings.

Stratifying by SOC strength (using maximum atomic number $Z_{\max}$ as a proxy), we observe a monotonic degradation from weak-SOC to strong-SOC regimes for both models (Table~\ref{tab:soc_regime}). For example, HamGNN increases from 1.10\,meV to 1.74\,meV and DeepH-pack from 1.90\,meV to 2.76\,meV, suggesting an intrinsic difficulty in learning relativistic effects rather than an architecture-specific artifact.
Element-level analysis further shows that heavy $5d$/$4f$ elements (e.g., Os, W, Pt, Lu) are systematically harder, whereas light main-group elements approach the numerical precision floor. At the structure level, 83.4\% of samples are well-predicted by both models (MAE $<$ 3\,meV), while the rare hard cases are concentrated in strong-SOC, small-cell, high-symmetry phases. Collectively, these results demonstrate that UniHam exposes SOC-driven and element-dependent failure modes that are invisible in light-element benchmarks, providing a more stringent testbed for Hamiltonian learning (see Appendix~\ref{app:detailed_experiments} for detailed per-element breakdowns and additional figures).

\begin{table}[h]
    \centering
    \caption{Hamiltonian MAE (meV) stratified by SOC strength regime, defined by the maximum atomic number $Z_{\max}$ in each structure.}
    \label{tab:soc_regime}
    \small
    \begin{tabular}{l r cc}
        \toprule
        \textbf{SOC Regime} & \textbf{N} & \textbf{HamGNN} & \textbf{DeepH-pack} \\
        \midrule
        Weak ($Z \leq 20$)                    & 200 & 1.10 & 1.90 \\
        Light-3$d$ ($21 \leq Z \leq 36$)      & 401 & 1.42 & 2.17 \\
        Medium-4$d$ ($37 \leq Z \leq 54$)     & 614 & 1.38 & 2.38 \\
        Strong-5$d$/4$f$ ($Z \geq 55$)        & 785 & 1.74 & 2.76 \\
        \bottomrule
    \end{tabular}
\end{table}

 \subsection{Out-of-Distribution Generalization}
 \label{sec:ood}

 We evaluate two OOD settings: system-size generalization (\textsc{OOD-Size}) and compositional generalization (\textsc{OOD-Comp}). For \textsc{OOD-Size}, both models exhibit near-perfect transfer from training cells ($\leq$60 atoms) to much larger structures (up to 444 atoms), with HamGNN achieving an OOD/ID MAE ratio of 0.89 and DeepH-pack showing a qualitatively similar trend, confirming that equivariant message-passing architectures inherit a strong locality inductive bias and generalize seamlessly across system sizes. In contrast, compositional OOD remains challenging and is highly architecture-dependent (Table~\ref{tab:ood_performance} and Appendix~\ref{app:detailed_experiments}). When entire element groups are removed from training, DeepH-pack degrades by $6$--$8\times$ relative to its ID reference, whereas HamGNN degrades by only $1.9$--$4.1\times$ and maintains single-digit-meV MAE across all three scenarios (In, RE, NG). Notably, the inter-model performance gap widens substantially under OOD conditions, demonstrating that architectural choices that yield modest gains in-distribution become decisive for zero-shot compositional generalization.

\begin{table}[t]
    \centering
    \small
    \setlength{\tabcolsep}{3pt}
    \caption{Compositional out-of-distribution generalization: Hamiltonian MAE (meV). ``ID'' rows report the MAE on the same element subgroups under the standard random-split model (elements present in training). ``OOD'' rows report the MAE when these elements are \emph{entirely removed} from training. Degradation factors ($\times$) are computed per element group.}
    \label{tab:ood_performance}
    \begin{tabular}{l l ccc}
        \toprule
        \textbf{Model} & \textbf{Setting} & \makecell[c]{\textbf{In}}& \makecell[c]{\textbf{RE}} & \makecell[c]{\textbf{NG}} \\
        \midrule
        \multirow{2}{*}{DeepH-pack}
            & ID   & 2.89 & 3.14 & 5.09 \\
            & OOD          & 22.49  & 22.73  & 31.48 \\
        \midrule
        \multirow{2}{*}{HamGNN}
            & ID   & 1.40 & 2.04 & 3.06 \\
            & OOD          & 5.71   & 5.39  & 5.93 \\
        \bottomrule
    \end{tabular}
\end{table}

\begin{table}[t]
    \centering
    \small
    \setlength{\tabcolsep}{4pt}
    \caption{Per-element OOD MAE (meV) on the rare-earth test set, sorted by atomic number. $N$: number of structures containing each element. Total MAE is computed over all real and imaginary Hamiltonian matrix elements.}
    \label{tab:ood_re_per_elem}
    \begin{tabular}{l c c c}
        \toprule
        \textbf{Elem.} & $N$ & \textbf{DeepH-pack} & \textbf{HamGNN} \\
        \midrule
        Pr ($4f^{2}$)   & 107 & 19.59 & 6.80 \\
        Nd ($4f^{3}$)   & 228 & 21.96 & 5.96 \\
        Sm ($4f^{5}$)   & 262 & 22.36 & 4.42 \\
        Dy ($4f^{10}$)  & 121 & 24.21 & 5.25 \\
        Ho ($4f^{11}$)  & 127 & 26.15 & 5.35 \\
        \bottomrule
    \end{tabular}
\end{table}

\subsection{Downstream Property Prediction}
\begin{figure*}[tbp]
    \centering
    \Description{Comparison of predicted and DFT-calculated band structures for multiple material systems.}
    \includegraphics[width=0.67\textwidth]{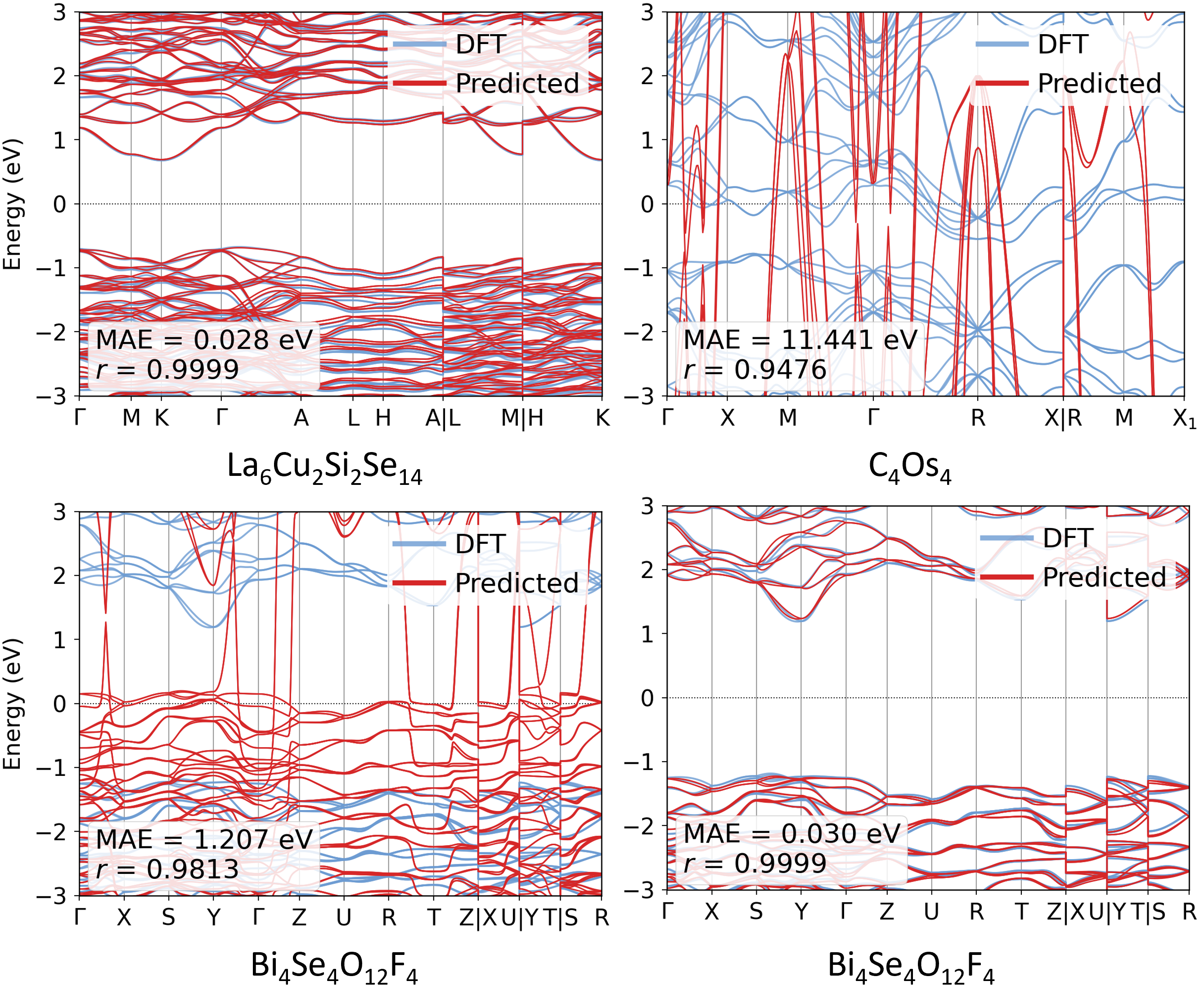}
    \caption{Qualitative comparison of predicted band structures. (Top left) La$_6$Cu$_2$Si$_2$Se$_{14}$ predicted by HamGNN (Base), achieving near-perfect agreement (band MAE = 0.028\,eV). (Top right) C$_4$Os$_4$ predicted by HamGNN (Base), showing severe deviations (band MAE = 11.4\,eV) due to strong SOC from Os. (Bottom) Bi$_4$Se$_4$O$_{12}$F$_4$ predicted by HamGNN Base (left) vs.\ HamGNN FT (right), demonstrating how eigenvalue-level fine-tuning corrects conduction-band artifacts.}
    \label{fig:band}
\end{figure*}

\begin{table*}[t]
\centering
\caption{Downstream property prediction accuracy on the test set. Band gap error is evaluated on 1{,}252 non-metallic structures; band structure and DOS metrics are computed on all 2{,}000 test structures within a $\pm$3\,eV window around the Fermi level.}
\label{tab:downstream_metrics}

\begin{adjustbox}{max width=\textwidth}
\begin{tabular}{l c c c c c}
\toprule
\multirow{2}{*}{\textbf{Model}} &
\multicolumn{1}{c}{\textbf{Band Gap}} &
\multicolumn{2}{c}{\textbf{Band Structure}} &
\multicolumn{2}{c}{\textbf{DOS}} \\
\cmidrule(lr){2-2}
\cmidrule(lr){3-4}
\cmidrule(lr){5-6}
& \textbf{MAE (eV)}
& \textbf{Pearson $r$}
& \textbf{MAE (eV)}
& \textbf{Pearson $r$}
& \textbf{MAE (st/eV)} \\
\midrule
DeepH-pack
& 1.925 & 0.893 & 1.795 & 0.088 & 8.29 \\
HamGNN (Base)
& 1.697 & 0.908 & 1.612 & 0.203 & 7.48 \\
HamGNN (FT)
& \textbf{0.622} & \textbf{0.988} & \textbf{0.360} & \textbf{0.555} & \textbf{4.73} \\
\bottomrule
\end{tabular}
\end{adjustbox}

\end{table*}

We evaluate the \textsc{Prop-id} task using band gap, band structure, and DOS metrics (Table~\ref{tab:downstream_metrics}). A clear hierarchy emerges. First, incorporating spectral-level supervision is crucial: HamGNN~(FT) dramatically outperforms its base counterpart, reducing band gap MAE from 1.697 to 0.622\,eV, band structure MAE from 1.612 to 0.360\,eV, and DOS MAE from 7.48 to 4.73\,states/eV, while increasing band Pearson correlation to 0.988. Second, Hamiltonian MAE alone is an insufficient proxy for downstream fidelity: although HamGNN~(Base) achieves substantially lower Hamiltonian MAE than DeepH-pack, this advantage translates only weakly to band gaps and band structures, motivating UniHam's inclusion of spectral-level evaluation. Finally, DOS reconstruction remains substantially harder than band structure prediction, with only moderate correlations even for HamGNN~(FT), highlighting an open challenge in accurately capturing fine-grained spectral distributions.

Figure~\ref{fig:band} provides qualitative illustration of these findings. The top row contrasts two HamGNN~(Base) predictions: La$_6$Cu$_2$Si$_2$Se$_{14}$ (left) achieves near-perfect band agreement (band MAE\,=\,0.028\,eV), as its constituent elements---La, Cu, Si, Se---are well-represented in training and belong to moderate SOC regimes. In contrast, C$_4$Os$_4$ (right) exhibits severe deviations (band MAE\,=\,11.4\,eV): Osmium ($Z{=}76$) is the heaviest $5d$ element in UniHam and the most challenging for both models (cf.\ Section~\ref{sec:experiments}), with strong SOC and limited training coverage (${\sim}$0.5\% of structures), causing Hamiltonian errors that are amplified through diagonalization. The bottom row demonstrates the impact of fine-tuning on Bi$_4$Se$_4$O$_{12}$F$_4$, a heavy-element system containing Bi ($Z{=}83$): HamGNN~(Base) (left) captures the overall band topology but produces noticeable deviations in the conduction manifold, while HamGNN~(FT) (right) largely eliminates these artifacts, confirming that eigenvalue-level supervision is critical for translating Hamiltonian accuracy into reliable spectral properties. See Appendix~\ref{app:detailed_experiments} for extended downstream analysis.

\subsection{Direct-Gap Materials Discovery}
\label{sec:direct_gap}

Beyond evaluating spectral fidelity on a fixed test set, we demonstrate that predicted Hamiltonians can enable practical materials screening. Direct-gap semiconductors are essential for optoelectronic applications (LEDs, lasers, solar cells) because their band-edge alignment at the same crystal momentum allows radiative recombination without phonon mediation. Identifying such materials conventionally requires full DFT+SOC calculations per candidate---a prohibitive bottleneck for high-throughput screening.

We use the UniHam-trained HamGNN~(FT) model to screen the 2{,}000-structure \textsc{Prop-id} test set. The model identifies 1{,}241 insulators (defined as having a predicted band gap $>$ 0.1\,eV). Among these, the predicted direct- vs.\ indirect-gap classification achieves \textbf{83.4\% consistency} with DFT ground truth, demonstrating that Hamiltonian-level predictions can recover the gap-type label that directly determines a material's technological relevance. Crucially, no DFT calculation is required at inference time: both the Hamiltonian prediction and the subsequent diagonalization to obtain band structures and gap types are purely model-driven.

These results establish that the UniHam benchmark, together with spectral-level supervision, produces models that are not merely accurate in the aggregate metric sense, but also actionable for downstream discovery tasks.

\subsection{Cross-Dataset Generalization on GNoME}
\label{sec:gnome}

A key test of a model's practical utility is whether it can generalize to structures from sources unseen during training, under realistic screening conditions. We evaluate this by applying the UniHam-trained HamGNN~(FT) model directly---without any retraining or fine-tuning---to \textbf{3,000 crystal structures} randomly sampled from the GNoME database~\cite{merchant2023scaling} with proportional sampling matching the Materials Project direct:indirect gap ratio.

The model screens 720 of the 3,000 structures as direct-gap semiconductors. We validate these predictions against the Materials Project database, which provides independent DFT reference calculations (with different pseudopotentials and exchange--correlation functionals).

\begin{table}[t]
\centering
\small
\caption{Band-gap distribution of 720 direct-gap candidates identified by HamGNN~(FT) on 3,000 GNoME structures.}
\label{tab:gnome_bandgap}
\setlength{\tabcolsep}{3pt}
\begin{tabular}{l c c c c c c c}
\toprule
Band gap (eV) & 0--1 & 1--2 & 2--3 & 3--4 & 4--5 & 5--6 & 6+ \\
\midrule
Count & 267 & 216 & 124 & 65 & 31 & 12 & 5 \\
Fraction (\%) & 37.1 & 30.0 & 17.2 & 9.0 & 4.3 & 1.7 & 0.7 \\
\bottomrule
\end{tabular}
\end{table}

\noindent\textbf{Results.} Among the 720 materials screened as direct-gap, \textbf{92\% are confirmed} in Materials Project. Table~\ref{tab:gnome_bandgap} presents the band-gap distribution of the predicted direct-gap candidates, which peaks in the 0--2\,eV range (67.1\%), consistent with the known prevalence of narrow-gap semiconductors in inorganic crystal databases. This cross-dataset generalization experiment demonstrates that an SOC-complete model trained on UniHam transfers effectively to an independently constructed dataset, providing an end-to-end demonstration of high-throughput materials screening without DFT at inference time.


\section{Conclusion}
We introduced UniHam, a large-scale, SOC-complete Hamiltonian dataset and benchmark comprising 100,000+ complex-valued Hermitian Hamiltonians across 72 elements and diverse crystal symmetries, addressing a critical data bottleneck for foundation models in quantum materials. Systematic evaluation of two representative equivariant architectures shows that while sub-meV Hamiltonian reconstruction is achievable in weak-SOC regimes, both models exhibit consistent degradation for heavy elements and strong SOC, revealing an intrinsic challenge in learning relativistic spin–orbital interactions. We further demonstrate near-perfect system-size transferability, but substantial differences in compositional OOD robustness, with HamGNN achieving markedly better zero-shot generalization than DeepH-pack. Importantly, we show that Hamiltonian MAE alone is an insufficient proxy for physical fidelity: incorporating spectral-level supervision dramatically improves band gaps, band structures, and DOS, highlighting the necessity of downstream-aware training objectives. Overall, UniHam establishes a reproducible, multidimensional benchmark that sharpens model comparisons and provides a foundation for developing next-generation, physically grounded Hamiltonian-learning models.

A key limitation of UniHam is the substantial computational and storage overhead inherent to Hamiltonian-level learning. Complex-valued Hamiltonians scale quadratically with the number of atomic orbitals, leading to large per-sample file sizes and a total dataset footprint that is orders of magnitude larger than typical scalar-property datasets. Correspondingly, training requires holding high-dimensional matrix targets and intermediate representations in memory, resulting in heavy GPU memory consumption and significantly slower training compared with energy/force or charge-density prediction. These costs currently constrain batch size, model capacity, and hyperparameter exploration. Future work will investigate more compact representations and memory-efficient training strategies to improve scalability.

\begin{acks}
 We sincerely thank all the anonymous reviewers for their valuable feedback. This research is supported by the National Natural Science Foundation of China (62461146204).
\end{acks}

\clearpage

\bibliographystyle{ACM-Reference-Format}
\bibliography{ref}

@article{li2022deep,
  title={Deep-learning density functional theory Hamiltonian for efficient ab initio electronic-structure calculation},
  author={Li, He and Wang, Zun and Zou, Nianlong and Ye, Meng and Xu, Runzhang and Gong, Xiaoxun and Duan, Wenhui and Xu, Yong},
  journal={Nature Computational Science},
  volume={2},
  number={6},
  pages={367--377},
  year={2022},
  publisher={Nature Publishing Group US New York}
}

@article{zhong2024universal,
  title={Universal machine learning Kohn--Sham Hamiltonian for materials},
  author={Zhong, Yang and Yu, Hongyu and Yang, Jihui and Guo, Xingyu and Xiang, Hongjun and Gong, Xingao},
  journal={Chinese Physics Letters},
  volume={41},
  number={7},
  pages={077103},
  year={2024},
  publisher={IOP Publishing}
}

@article{yin2025advancing,
  title={Advancing Universal Deep Learning for Electronic-Structure Hamiltonian Prediction of Materials},
  author={Yin, Shi and Dai, Zujian and Pan, Xinyang and He, Lixin},
  journal={arXiv preprint arXiv:2509.19877},
  year={2025}
}

@article{tang2025deep,
  title={Deep-learning electronic structure calculations},
  author={Tang, Zechen and Chen, Haoxiang and Li, Yang and Qian, Yubing and Wang, Yuxiang and Fu, Weizhong and Li, Jialin and Si, Chen and Duan, Wenhui and Chen, Ji and others},
  journal={Nature Computational Science},
  pages={1--14},
  year={2025},
  publisher={Nature Publishing Group}
}

@article{wang2024universal,
  title={Universal materials model of deep-learning density functional theory Hamiltonian},
  author={Wang, Yuxiang and Li, Yang and Tang, Zechen and Li, He and Yuan, Zilong and Tao, Honggeng and Zou, Nianlong and Bao, Ting and Liang, Xinghao and Chen, Zezhou and others},
  journal={Science Bulletin},
  volume={69},
  number={16},
  pages={2514--2521},
  year={2024},
  publisher={Elsevier}
}

@article{li2025critical,
  title={A critical review of machine learning interatomic potentials and Hamiltonian},
  author={Li, Yifan and Zhang, Xiuying and Shen, Lei},
  journal={Journal of Materials Informatics},
  volume={5},
  number={4},
  pages={N--A},
  year={2025},
  publisher={OAE Publishing Inc.}
}

@article{huang2023central,
  title={The central role of density functional theory in the AI age},
  author={Huang, Bing and von Rudorff, Guido Falk and von Lilienfeld, O Anatole},
  journal={Science},
  volume={381},
  number={6654},
  pages={170--175},
  year={2023},
  publisher={American Association for the Advancement of Science}
}

@article{gong2023general,
  title={General framework for E (3)-equivariant neural network representation of density functional theory Hamiltonian},
  author={Gong, Xiaoxun and Li, He and Zou, Nianlong and Xu, Runzhang and Duan, Wenhui and Xu, Yong},
  journal={Nature Communications},
  volume={14},
  number={1},
  pages={2848},
  year={2023},
  publisher={Nature Publishing Group UK London}
}

@article{jain2016computational,
  title={Computational predictions of energy materials using density functional theory},
  author={Jain, Anubhav and Shin, Yongwoo and Persson, Kristin A},
  journal={Nature Reviews Materials},
  volume={1},
  number={1},
  pages={1--13},
  year={2016},
  publisher={Nature Publishing Group}
}

@article{marzari2021electronic,
  title={Electronic-structure methods for materials design},
  author={Marzari, Nicola and Ferretti, Andrea and Wolverton, Chris},
  journal={Nature materials},
  volume={20},
  number={6},
  pages={736--749},
  year={2021},
  publisher={Nature Publishing Group UK London}
}

@article{galitski2013spin,
  title={Spin--orbit coupling in quantum gases},
  author={Galitski, Victor and Spielman, Ian B},
  journal={Nature},
  volume={494},
  number={7435},
  pages={49--54},
  year={2013},
  publisher={Nature Publishing Group UK London}
}

@article{prayitno2018implementation,
  title={Implementation of generalized Bloch theorem using linear combination of pseudo-atomic orbitals},
  author={Prayitno, Teguh Budi and Ishii, Fumiyuki},
  journal={Journal of the Physical Society of Japan},
  volume={87},
  number={11},
  pages={114709},
  year={2018},
  publisher={The Physical Society of Japan}
}

@article{chen2022improving,
  title={Improving material property prediction by leveraging the large-scale computational database and deep learning},
  author={Chen, Pin and Chen, Jianwen and Yan, Hui and Mo, Qing and Xu, Zexin and Liu, Jinyu and Zhang, Wenqing and Yang, Yuedong and Lu, Yutong},
  journal={The Journal of Physical Chemistry C},
  volume={126},
  number={38},
  pages={16297--16305},
  year={2022},
  publisher={ACS Publications}
}

@article{jain2013commentary,
  title={Commentary: The Materials Project: A materials genome approach to accelerating materials innovation},
  author={Jain, Anubhav and Ong, Shyue Ping and Hautier, Geoffroy and Chen, Wei and Richards, William Davidson and Dacek, Stephen and Cholia, Shreyas and Gunter, Dan and Skinner, David and Ceder, Gerbrand and others},
  journal={APL materials},
  volume={1},
  number={1},
  year={2013},
  publisher={AIP Publishing}
}

@article{hohenberg1964inhomogeneous,
  title={Inhomogeneous electron gas},
  author={Hohenberg, Pierre and Kohn, Walter},
  journal={Physical Review},
  volume={136},
  number={3B},
  pages={B864},
  year={1964}
}

@article{kohn1965self,
  title={Self-consistent equations including exchange and correlation effects},
  author={Kohn, Walter and Sham, Lu Jeu},
  journal={Physical Review},
  volume={140},
  number={4A},
  pages={A1133},
  year={1965}
}

@article{perdew1996generalized,
  title={Generalized gradient approximation made simple},
  author={Perdew, John P and Burke, Kieron and Ernzerhof, Matthias},
  journal={Physical Review Letters},
  volume={77},
  number={18},
  pages={3865},
  year={1996}
}

@article{ozaki2003variationally,
  title={Variationally optimized atomic orbitals for large-scale electronic structures},
  author={Ozaki, Taisuke},
  journal={Physical Review B},
  volume={67},
  pages={155108},
  year={2003}
}

@article{ozaki2004numerical,
  title={Numerical atomic basis orbitals from {H} to {Kr}},
  author={Ozaki, Taisuke and Kino, Hiori},
  journal={Physical Review B},
  volume={69},
  pages={195113},
  year={2004}
}

@article{dudarev1998electron,
  title={Electron-energy-loss spectra and the structural stability of nickel oxide: An {LSDA+U} study},
  author={Dudarev, S L and Botton, G A and Savrasov, S Y and Humphreys, C J and Sutton, A P},
  journal={Physical Review B},
  volume={57},
  number={3},
  pages={1505},
  year={1998}
}

@article{graulis2012crystallography,
  title={Crystallography {Open Database (COD)}: an open-access collection of crystal structures and platform for world-wide collaboration},
  author={Graulis, Saulius and Chateigner, Daniel and Downs, Robert T and Yokochi, A F T and Quir{\'o}s, Miguel and Lutterotti, Luca and Manakova, Elena and Butkus, Juozas and Moeck, Peter and Le Bail, Armel},
  journal={Journal of Applied Crystallography},
  volume={42},
  pages={726--729},
  year={2009}
}

@article{belsky2002icsd,
  title={New developments in the {Inorganic Crystal Structure Database (ICSD)}: accessibility in support of materials research and design},
  author={Belsky, Alec and Hellenbrandt, Mariette and Karen, Vicky Lynn and Luksch, Peter},
  journal={Acta Crystallographica Section B},
  volume={58},
  pages={364--369},
  year={2002}
}

@article{geiger2022e3nn,
  title={e3nn: {Euclidean} Neural Networks},
  author={Geiger, Mario and Smidt, Tess E},
  journal={arXiv preprint arXiv:2207.09453},
  year={2022}
}

@inproceedings{thomas2018tensor,
  title={Tensor field networks: Rotation- and translation-equivariant neural networks for {3D} point clouds},
  author={Thomas, Nathaniel and Smidt, Tess and Kearnes, Steven and Yang, Lusann and Li, Li and Kober, Kai and Riley, Patrick},
  booktitle={arXiv preprint arXiv:1802.08219},
  year={2018}
}

@article{hegde2017machine,
  title={Machine-learned approximations to density functional theory {Hamiltonians}},
  author={Hegde, Ganesh and Bowen, R Chris},
  journal={Scientific Reports},
  volume={7},
  pages={42669},
  year={2017}
}

@article{schutt2019unifying,
  title={Unifying machine learning and quantum chemistry with a deep neural network for molecular wavefunctions},
  author={Sch{\"u}tt, Kristof T and Gastegger, Michael and Tkatchenko, Alexandre and M{\"u}ller, Klaus-Robert and Maurer, Reinhard J},
  journal={Nature Communications},
  volume={10},
  pages={5024},
  year={2019}
}

@article{zhong2023transferable,
  title={Transferable equivariant graph neural networks for the {Hamiltonians} of molecules and solids},
  author={Zhong, Yang and Yu, Hongyi and Su, Mao and Gong, Xingao and Xiang, Hongjun},
  journal={npj Computational Materials},
  volume={9},
  number={1},
  pages={182},
  year={2023}
}

@article{li2023deep_magnetic,
  title={Deep-learning electronic-structure calculation of magnetic superstructures},
  author={Li, He and Tang, Zun and Gong, Xiaoxun and Zou, Nianlong and Duan, Wenhui and Xu, Yong},
  journal={Nature Computational Science},
  volume={3},
  pages={321},
  year={2023}
}

@article{merchant2023scaling,
  title={Scaling deep learning for materials discovery},
  author={Merchant, Amil and Batzner, Simon and Schoenholz, Samuel S and Aykol, Muratahan and Cheon, Gowoon and Cubuk, Ekin Dogus},
  journal={Nature},
  volume={624},
  pages={80--85},
  year={2023}
}

@article{ong2013python,
  title={Python {Materials Genomics} ({pymatgen}): A robust, open-source python library for materials analysis},
  author={Ong, Shyue Ping and Richards, William Davidson and Jain, Anubhav and Hautier, Geoffroy and Kocher, Michael and Cholia, Shreyas and Gunter, Dan and Chevrier, Vincent L and Persson, Kristin A and Ceder, Gerbrand},
  journal={Computational Materials Science},
  volume={68},
  pages={314--319},
  year={2013}
}

@article{zhang2018deep,
  title={Deep potential molecular dynamics: A scalable model with the accuracy of quantum mechanics},
  author={Zhang, Linfeng and Han, Jiequn and Wang, Han and Car, Roberto and E, Weinan},
  journal={Physical Review Letters},
  volume={120},
  pages={143001},
  year={2018}
}

@article{behler2007generalized,
  title={Generalized neural-network representation of high-dimensional potential-energy surfaces},
  author={Behler, J{\"o}rg and Parrinello, Michele},
  journal={Physical Review Letters},
  volume={98},
  pages={146401},
  year={2007}
}

\appendix

\section{Dataset Construction and Statistics}
\label{app:data_generation}

\begin{figure}[ht]
    \centering
    \Description{Flowchart of the UniHam dataset construction pipeline.}
    \includegraphics[width=\columnwidth]{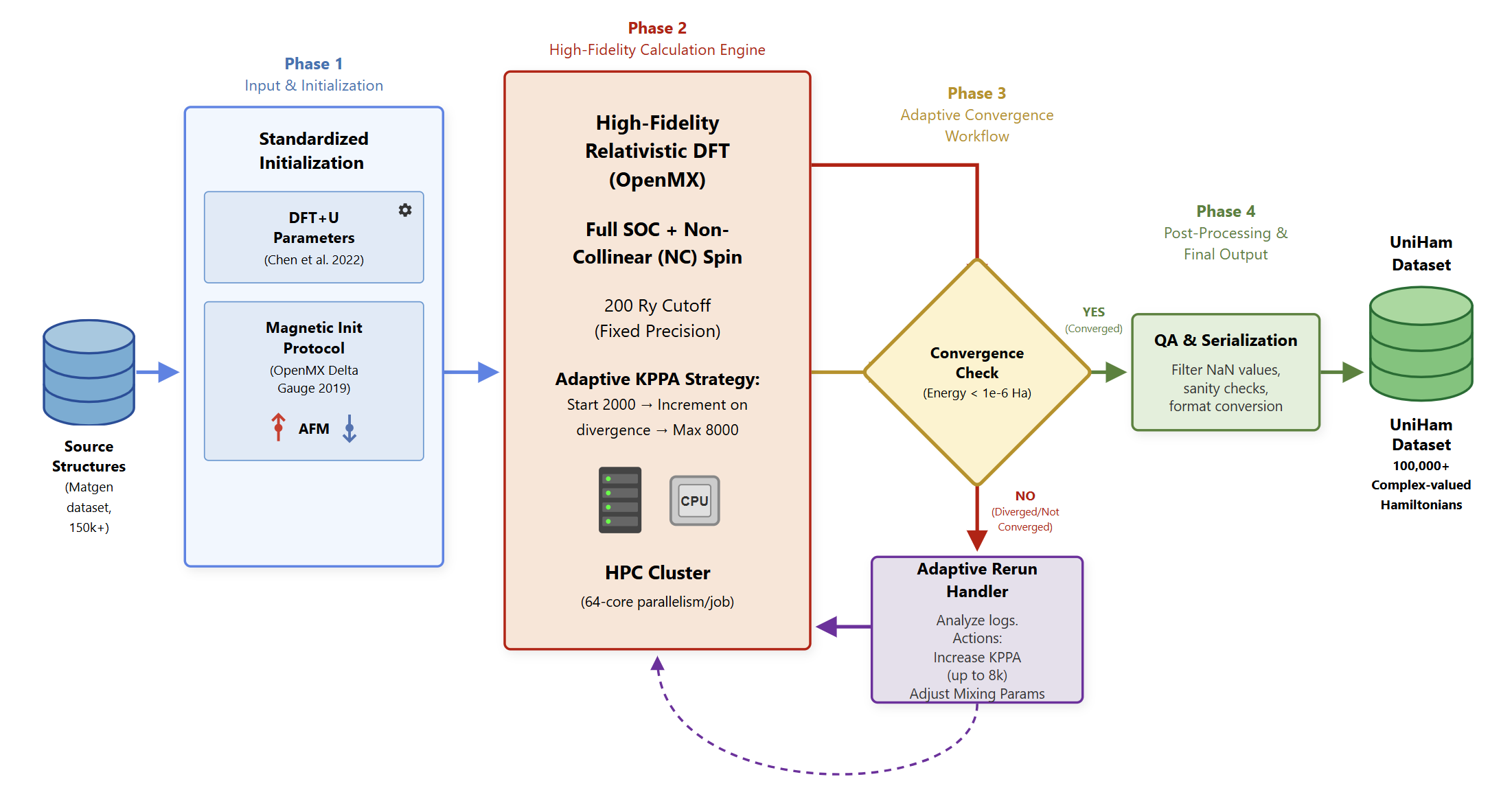}
    \caption{UniHam Dataset Construction Pipeline.}
    \label{fig:data_pipeline}
\end{figure}

We developed a fault-tolerant Python-based pipeline using Pymatgen~\cite{ong2013python} to manage over 150,000 DFT calculations. The pipeline (Figure~\ref{fig:data_pipeline}) consists of three stages: (1) automated generation of OpenMX input files with element-specific basis sets, pseudopotentials, and DFT+U parameters; (2) distributed execution across HPC clusters; (3) adaptive SCF convergence with automated rerun on failure. $\sim$33\% of initial candidates were filtered, yielding 103,121 verified entries. All calculations used OpenMX 3.9~\cite{ozaki2003variationally} under PBE~\cite{perdew1996generalized} with non-collinear spin and full SOC, 350\,Ry cutoff, $10^{-6}$\,Ha SCF convergence, $\geq$2000\,kppa $k$-point density, and DFT+U corrections for 30 elements~\cite{dudarev1998electron}. The dataset spans 72 elements, 225/230 space groups, 1--444 atoms/cell (mean 30), with 66.6\% containing heavy elements ($Z{\ge}37$). Data sources: ICSD (43\%), Matgen (53\%), COD (5\%).

\begin{table}[h]
\centering
\small
\caption{Summary statistics of the UniHam dataset.}
\label{tab:dataset-stats}
\setlength{\tabcolsep}{3pt}
\begin{tabular}{lc}
\toprule
\textbf{Property} & \textbf{Value} \\
\midrule
Total structures & 103,121 \\
Unique compositions & 50,077 \\
Elements covered & 72 / Heavy ($Z{\ge}37$): 41 \\
Structures with heavy elements & 66.6\% \\
Atoms per cell (mean/median/range) & 30 / 20 / 1--444 \\
Space groups covered & 225 / 230 \\
Crystal systems covered & 7 / 7 \\
\bottomrule
\end{tabular}
\end{table}

\section{Implementation Details}
\label{app:impl_details}

The training set is restricted to structures with $\leq$60 atoms per unit cell because SOC-complete Hamiltonians scale quadratically with the number of spinor orbitals: for a structure with $N$ atoms, each atom carries up to 52 spinor components (for the \texttt{s3p2d2f1} basis), and the full Hamiltonian matrix dimension is $O(N^2)$. Even on NVIDIA H100 GPUs (80\,GB) with a batch size of 1, structures exceeding $\sim$60 atoms cause out-of-memory failures during back-propagation. The remaining 4{,}328 structures with $>$60 atoms are reserved for the OOD-Size evaluation. Table~\ref{tab:hyperparams} lists the hyperparameters.

\begin{table}[h]
\centering
\small
\caption{Hyperparameters for baseline models.}
\label{tab:hyperparams}
\setlength{\tabcolsep}{2pt}
\begin{tabular}{lcc}
\toprule
\textbf{Hyperparameter} & \textbf{DeepH-pack} & \textbf{HamGNN} \\
\midrule
Architecture & DeepH-E3 (Eagle) & HamGNN \\
Message-passing layers & 3 & 3 \\
Radial cutoff & 10.0\,\AA & 26.0\,Bohr \\
Radial basis & Gaussian & 64 Bessel \\
Attention heads & 2 & --- \\
Delta learning ($H_0$) & No & Yes \\
Hermitian constraint & No & Yes \\
\midrule
Optimizer & AdamW & Adam \\
Initial learning rate & $10^{-3}$ & $10^{-3}$ \\
LR scheduler & ReduceOnPlateau & ReduceOnPlateau \\
Weight decay & $10^{-3}$ & --- \\
Gradient clipping & --- & 1.0 \\
Batch size & 1 & 1 ($\times$4 accum.) \\
GPUs & 8$\times$ H100 & 8$\times$ H100 \\
\bottomrule
\end{tabular}
\end{table}

DeepH-pack uses epoch-level data sharding to handle the 86k training set (partitioning into $K$ disjoint shards $\leq$10k each). HamGNN uses gradient accumulation (4 steps) for 100 epochs; HamGNN~(FT) adds band-energy loss over 5 random $k$-points (8 bands near VBM) from the converged Base checkpoint.

\section{Extended Experimental Results}
\label{app:detailed_experiments}

\begin{figure}[ht]
    \centering
    \Description{Scatter plot comparing per-structure MAE between HamGNN and DeepH-pack models.}
    \includegraphics[width=\columnwidth]{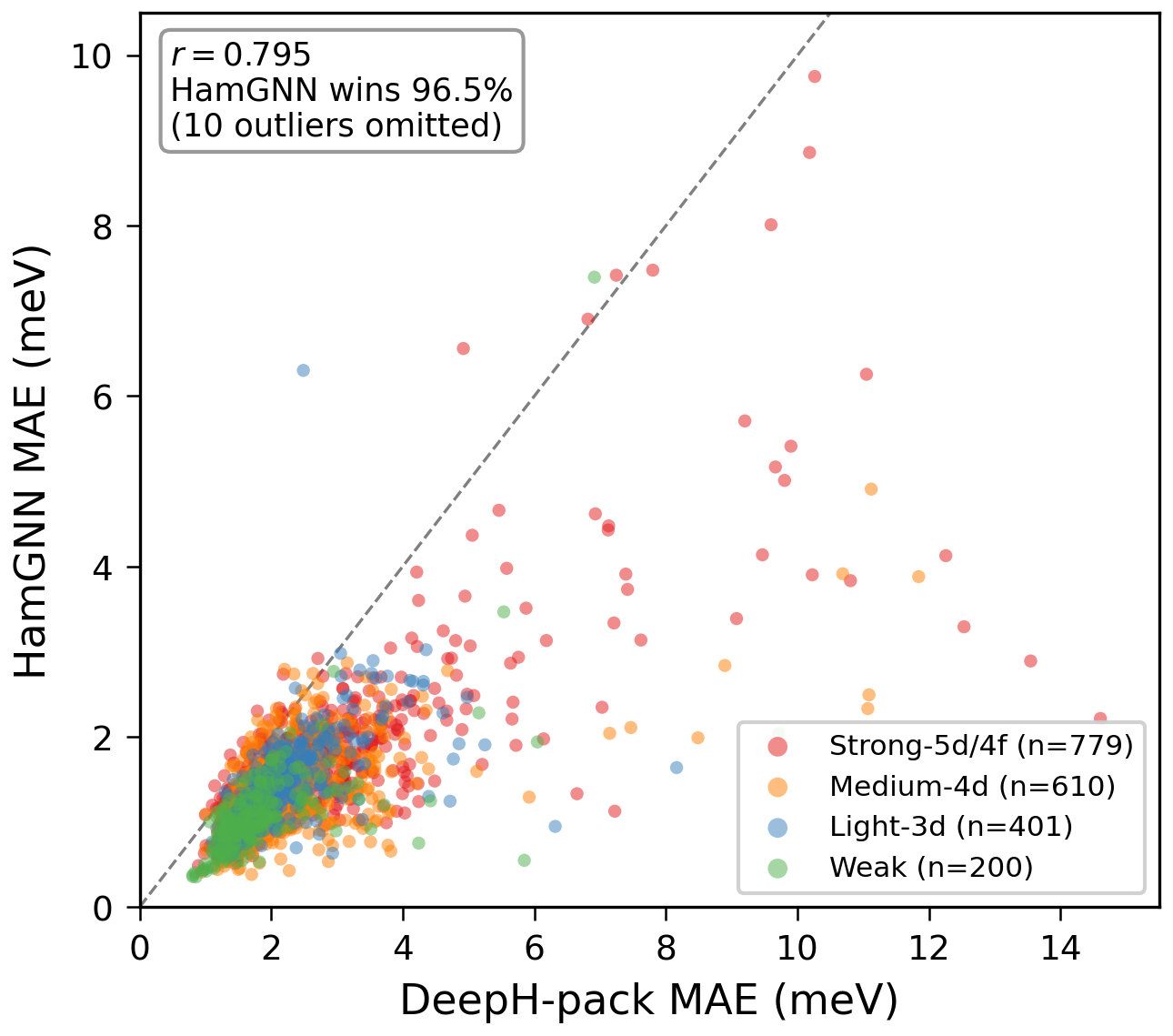}
    \caption{Per-structure MAE comparison between HamGNN and DeepH-pack on the test set. Each point represents one structure; the dashed line indicates equal performance. Pearson $r=0.795$.}
    \label{fig:scatter}
\end{figure}

\textbf{Model comparison.} HamGNN achieves a total MAE of 1.50\,meV, outperforming DeepH-pack (2.44\,meV) by $1.6\times$. HamGNN attains lower MAE on 96.5\% of individual test structures (1{,}931 out of 2{,}000). The Pearson correlation between per-structure MAEs of the two models is $r=0.795$ (Figure~\ref{fig:scatter}), indicating that both models find a largely overlapping set of structures challenging.

\textbf{Real and imaginary error decomposition.} The imaginary part yields substantially smaller absolute errors for both models (HamGNN: 0.041\,meV; DeepH-pack: 0.385\,meV) compared to the real part (HamGNN: 2.96\,meV; DeepH-pack: 4.49\,meV). The imaginary-to-real MAE ratio differs markedly between models: approximately 1.4\% for HamGNN versus 8.6\% for DeepH-pack. HamGNN's real-part MAE grows by $1.57\times$ from the weak to the strong SOC regime (2.18$\to$3.42\,meV), while imaginary-part MAE grows by $2.38\times$ (0.021$\to$0.049\,meV). Although imaginary errors remain small in absolute terms, their disproportionate growth with SOC strength indicates that spin-flip interaction terms represent the frontier challenge in Hamiltonian learning.

\begin{figure}[ht]
    \centering
    \Description{Bar chart of Hamiltonian MAE stratified by SOC strength regime.}
    \includegraphics[width=\columnwidth]{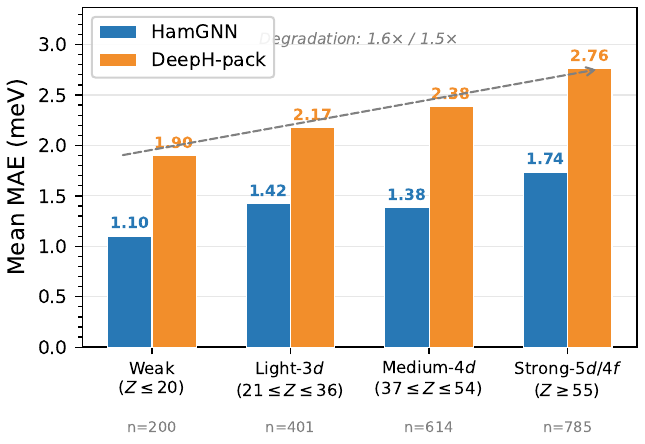}
    \caption{Hamiltonian MAE (meV) stratified by SOC strength regime for both models.}
    \label{fig:soc_bar}
\end{figure}

\textbf{SOC-stratified analysis.} Figure~\ref{fig:soc_bar} visualizes the SOC-stratified results (Table~\ref{tab:soc_regime} in the main text). Both models exhibit a clear monotonic increase in MAE from the weak-SOC to the strong-SOC regime. HamGNN's mean MAE rises by 58\% (1.10$\to$1.74\,meV) and DeepH-pack's by 45\% (1.90$\to$2.76\,meV).

\textbf{Crystal system variation.} MAE varies across crystal systems, with orthorhombic structures being the hardest (DeepH-pack: 2.84\,meV) and monoclinic the easiest (2.08\,meV). \textbf{Atom count effects.} Small unit cells (1--4 atoms) are counterintuitively harder than large ones, likely due to reduced statistical averaging over matrix elements.

\begin{figure}[ht]
    \centering
    \Description{Periodic table heatmap of element-wise mean Hamiltonian MAE.}
    \includegraphics[width=\columnwidth]{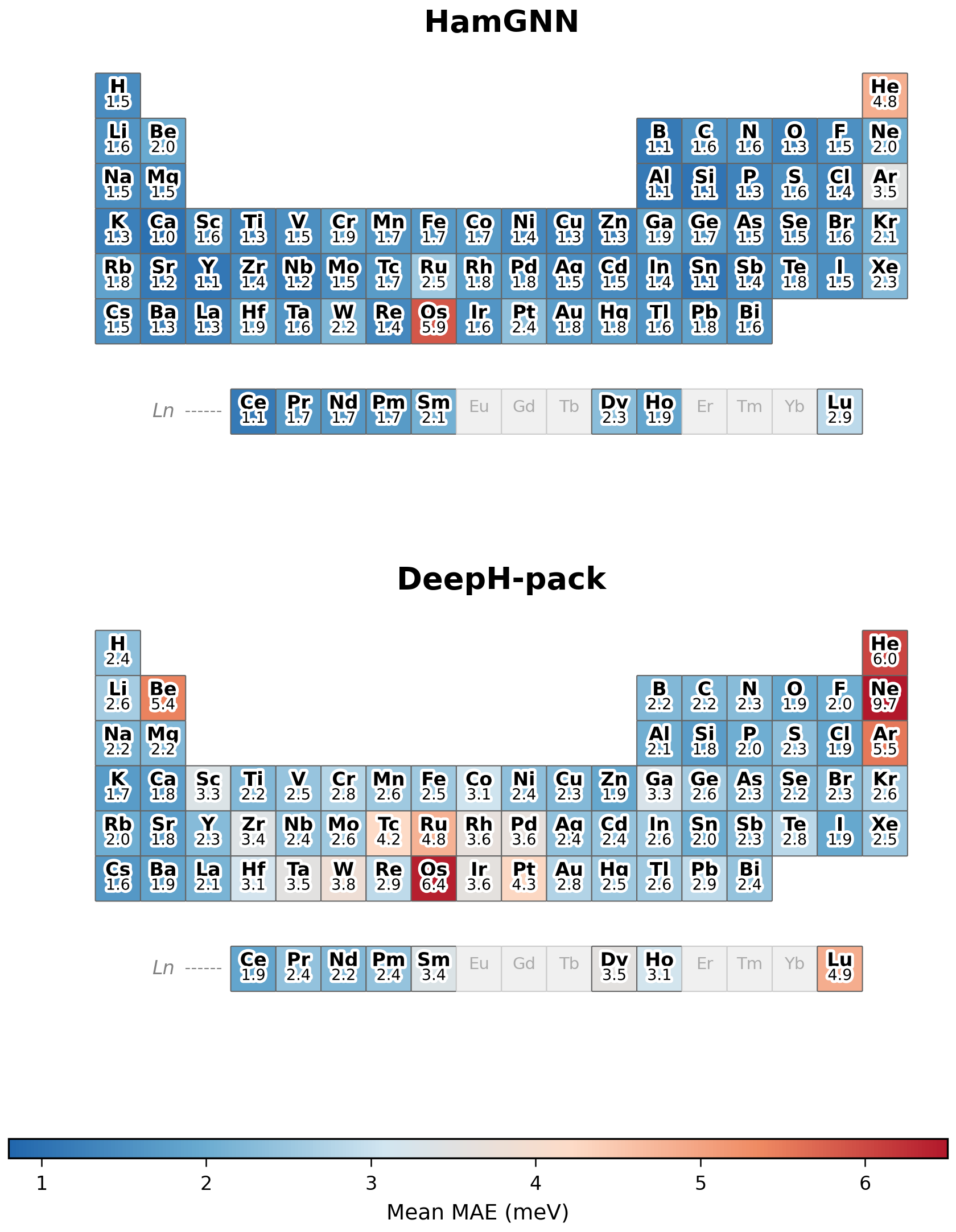}
    \caption{Element-wise mean Hamiltonian MAE (meV) mapped onto the periodic table.}
    \label{fig:periodic_table}
\end{figure}

\textbf{Element-level and structural difficulty.} Figure~\ref{fig:periodic_table} maps the mean MAE 
of each element onto the periodic table. The visualization reveals a clear gradient: heavier elements 
are systematically harder to predict. Osmium (Os, $Z{=}76$) is the most challenging element for both 
models (HamGNN: 5.87\,meV; DeepH-pack: 6.42\,meV), exceeding the overall average by $3$--$4\times$. 
More broadly, the 5$d$ transition metals---Ta, W, Os, Ir, and Pt---consistently produce elevated MAEs 
and exhibit the largest inter-model performance gaps ($>$1.5\,meV advantage for HamGNN), pointing to 
a regime where architectural differences matter most. At the other extreme, light main-group elements 
such as Si (HamGNN: 1.05\,meV; DeepH-pack: 1.76\,meV) and Ca (HamGNN: 0.98\,meV; DeepH-pack: 
1.76\,meV) approach the numerical precision floor. At the structure level, 83.4\% of structures are 
well-predicted by both models (MAE $<$3\,meV), while only 4 structures ($<$0.2\%) are simultaneously 
hard for both (MAE $>$10\,meV). All four belong to the strong-SOC regime, share the orthorhombic 
space group $Immm$ (No.~71) with just 4 atoms per cell, and contain $5d$ transition metals (W, Os, Pt). 
Conversely, the easiest structures are dominated by light-element oxides and silicates (e.g., SiO$_2$, 
CaO variants), achieving average MAEs below 0.7\,meV. These findings demonstrate that UniHam's broad 
elemental and structural coverage exposes meaningful performance variations that would remain invisible 
in datasets restricted to light elements.

\textbf{System-size OOD generalization.} HamGNN achieves an OOD/ID MAE ratio of 0.89 (OOD 1.34\,meV vs.\ ID 1.50\,meV), demonstrating near-perfect size transferability from $\leq$60-atom training to 444-atom test structures. DeepH-pack exhibits a qualitatively similar pattern. This is physically intuitive: equivariant message-passing architectures operate on local atomic neighborhoods defined by a radial cutoff, so prediction for each matrix block depends only on the local chemical environment rather than the global system size.

\textbf{Compositional OOD: detailed findings.} Both models are configured with the same equivariant representation capacity, ensuring that performance differences arise from architectural design rather than angular-momentum truncation. Architecture is decisive for OOD robustness: DeepH-pack's MAE rises by 6--8$\times$ relative to its per-group ID reference, whereas HamGNN degrades by only 1.9--4.1$\times$. The inter-model gap widens under OOD conditions from 1.5--2.1$\times$ (ID) to 3.9--5.3$\times$ (OOD). HamGNN maintains single-digit-meV accuracy across all three OOD scenarios (5.39--5.93\,meV). Its OOD MAE on NG (5.93\,meV) is only $1.9\times$ its ID reference and is even lower than DeepH-pack's ID performance on the same structures (5.09\,meV). HamGNN achieves remarkably uniform OOD MAE across all three scenarios, while DeepH-pack exhibits pronounced dependence on chemical distance (NG: 31.48\,meV vs.\ In/RE: $\sim$22.5\,meV).

\textbf{Downstream property prediction: detailed analysis.} HamGNN~(FT) reduces band gap error by $2.7\times$ (1.697$\to$0.622\,eV) and band structure MAE by $4.5\times$ (1.612$\to$0.360\,eV). The median band gap error drops from 1.134 to 0.295\,eV. Critically, Hamiltonian MAE alone is an insufficient proxy: HamGNN~(Base) achieves a 38\% lower Hamiltonian MAE than DeepH-pack (1.50 vs.\ 2.44\,meV), yet this translates to only $\sim$10\% band structure MAE improvement (1.612 vs.\ 1.795\,eV). This directly justifies UniHam's multi-dimensional evaluation protocol. DOS reconstruction remains a frontier challenge: even HamGNN~(FT) achieves only moderate DOS correlation (0.555), in stark contrast to its near-perfect band structure correlation (0.988). The band derivative correlation follows a consistent pattern: 0.711 for HamGNN~(FT), 0.277 for HamGNN~(Base), and 0.176 for DeepH-pack, confirming that eigenvalue fine-tuning improves not only energy accuracy but also the topological features of the electronic structure.

\section{Data and Code Availability}
\label{sec:availability}

The UniHam dataset, benchmark splits, evaluation code, and model training scripts are publicly available at \url{https://github.com/WendyYW/UniHam.git}. All 103,000+ structures with converged, complex-valued Hermitian Hamiltonians (full SOC) are downloadable via provided scripts with selective download support. Data formats are directly compatible with DeepH and HamGNN interfaces. We provide complete documentation for reproducing all experimental results.

\end{document}